\documentclass[letterpaper, 10 pt, conference]{ieeeconf}  

\IEEEoverridecommandlockouts                              

\usepackage{graphicx} 
\usepackage{amsmath} 
\usepackage{algorithm}
\usepackage{algpseudocode}
\usepackage{booktabs}
\usepackage{caption}
\usepackage{adjustbox}

\title{\LARGE \bf
Context-aware Risk Assessment and Its Application \\in Autonomous Driving
}

\author{Boyang Tian$^{1}$ and Weisong Shi$^{1}$
\thanks{$^{1}$ Department of Computer and Information Sciences, University of Delaware, Newark, DE 19716, USA
 {\tt\small \{tby,weisong\}@udel.edu}}
}

\begin{document}

\maketitle
\thispagestyle{empty}
\pagestyle{empty}

\begin{abstract}
Ensuring safety in autonomous driving requires precise, real-time risk assessment and adaptive behavior. Prior work on risk estimation either outputs coarse, global scene-level metrics lacking interpretability, proposes indicators without concrete integration into autonomous systems, or focuses narrowly on specific driving scenarios. We introduce the Context-aware Risk Index (CRI), a light-weight modular framework that quantifies directional risks based on object kinematics and spatial relationships, dynamically adjusting control commands in real time. CRI employs direction-aware spatial partitioning within a dynamic safety envelope using Responsibility-Sensitive Safety (RSS) principles, a hybrid probabilistic-max fusion strategy for risk aggregation, and an adaptive control policy for real-time behavior modulation. We evaluate CRI on the Bench2Drive benchmark comprising 220 safety-critical scenarios using a state-of-the-art end-to-end model Transfuser++ on challenging routes. Our collision-rate metrics show a 19\% reduction (p = 0.003) in vehicle collisions per failed route, a 20\% reduction (p = 0.004) in collisions per kilometer, a 17\% increase (p = 0.016) in composed driving score, and a statistically significant reduction in penalty scores (p = 0.013) with very low overhead (3.6 ms per decision cycle). These results demonstrate that CRI substantially improves safety and robustness in complex, risk-intensive environments while maintaining modularity and low runtime overhead.
\end{abstract}

\section{Introduction}
\label{sec:introduction}

Autonomous driving systems have made remarkable progress in recent years, yet ensuring safety and adaptability remains one of the most critical challenges in the field, particularly in dynamic and risk-intensive environments. As autonomous vehicles increasingly operate in complex real-world scenarios, the need for precise, real-time risk assessment becomes paramount. This paper addresses the problem of real-time, directional, and interpretable risk estimation for autonomous vehicles. Specifically, we ask: how can an autonomous system adapt its control strategies to a safer style on the fly by quantifying not just the presence of risk, but its direction, type, and severity? To address this challenge, we propose the Context-aware Risk Index (CRI), a lightweight, modular framework designed to enable real-time, risk-aware driving behavior adaptation.

Current approaches to risk estimation in autonomous driving face several fundamental limitations. Some of these works either output coarse, global scene-level metrics lacking interpretability \cite{yu2021complexity}, propose indicators without concrete integration into autonomous systems \cite{liu2019towards}, or focus narrowly on specific driving scenarios \cite{kutela2024influence, naveiro2024adversarial}. These limitations make it difficult to provide fine-grained, context-sensitive feedback that can be effectively incorporated into modern autonomous driving stacks.

Drawing from these observations, our solution is built upon three key insights. First, driving risk is inherently directional and object-specific; therefore, we generate per-object, direction-aware risk scores by analyzing each object's kinematic relationship to the ego vehicle making them highly explainable, rather than computing a single global value. Second, a usable risk model must be sufficiently generic to operate robustly across diverse driving scenarios, rather than being specialized for a limited subset of cases. Third, risk estimation is more meaningful if it actively influences system behavior. Accordingly, CRI is implemented as a lightweight module that interfaces directly with autonomous vehicle control subsystems, dynamically modulating the vehicle's driving style based on localized risk assessments.

To validate CRI's capability, we integrated it into a state-of-the-art end-to-end model Transfuser++ \cite{Jaeger_Chitta_Geiger_2023} and evaluated it on the Bench2Drive benchmark \cite{Jia_Yang_Li_Zhang_Yan_2024} in the CARLA simulator \cite{Dosovitskiy17}. Experiments show that CRI significantly reduces vehicle collisions, improves driving efficiency, and enhances comfort across both the full scenario set and a failure-prone subset, demonstrating safer and more reliable behavior under challenging conditions.

\begin{figure*}[htbp]
    \centering
    \includegraphics[width=1\textwidth]{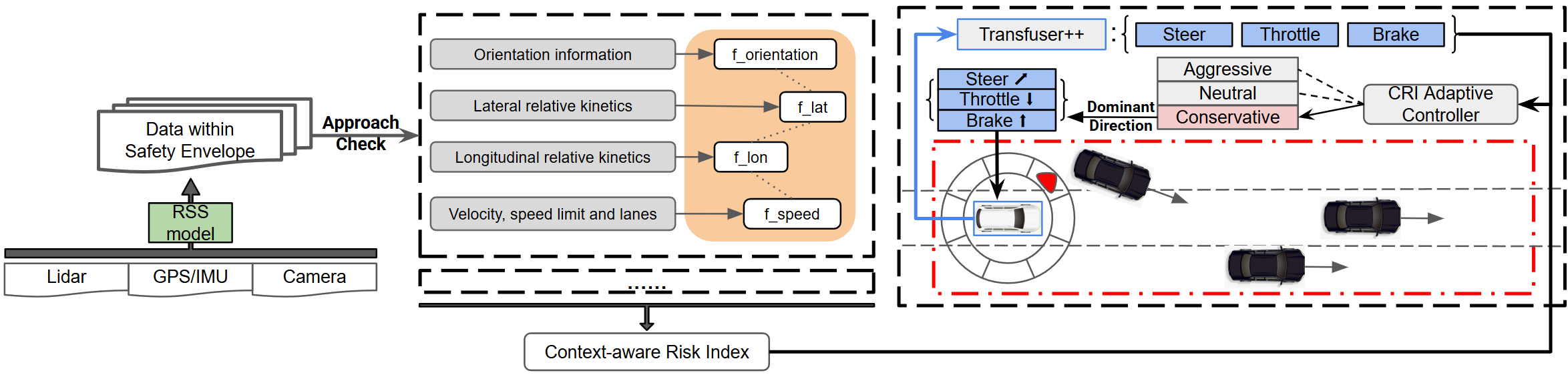}
    \caption{System overview of CRI integration.CRI computes directional risk from RSS-filtered objects and adjusts Transfuser++ control via aggregated value and dominant direction.}
    \label{fig:system_architecture}
\end{figure*}

\section{Related Works}
\label{sec:related}

Understanding and quantifying the complexity of driving environments is crucial for adapting autonomous vehicles (AVs) to diverse road conditions. Montewka {\em et al.} \cite{montewka2021collision} introduced a ship–ship collision criterion whose angle‐calculation is also pertinent to collision‐risk analyses in road contexts. Yu {\em et al.} \cite{yu2021complexity} proposed a complexity metric, but its output range and interpretability were limited. Other researchers explored complexity from the driver's perspective: Gold  {\em et al.} \cite{gold2016taking} examined how traffic density affects take‐over control from highly automated vehicles, while Faure  {\em et al.} \cite{faure2016effects} used subjective workload ratings that do not easily scale. Meanwhile, various data‐driven frameworks emerged: Guo  {\em et al.} \cite{guo2020safedata} reviewed open‐source driving datasets via a driveability metric, Liu and Hansen \cite{liu2019towards} combined OpenStreetMap data with dynamic vehicle information, and Wang  {\em et al.} \cite{wang2018traffic} performed scenario classification based on traffic data. Manawadu  {\em et al.} \cite{manawadu2018estimating} used a simulator to estimate driver workload in different traffic densities. However, despite these efforts, many either lack a real‐time focus on ego‐vehicle kinetics or omit explicit collision‐risk factors such as relative angle or overspeed. Feng  {\em et al.} \cite{feng2023dense} and Garefalakis  {\em et al.} \cite{garefalakis2022data} use deep reinforcement learning and imbalanced classification for risk assessment, but lack explainability and direct integration with control adaptation. Consequently, a more kinetic‐aware and interpretable index—integrating multiple dimensions—remains needed.
\vspace{-2pt}  

Another line of work emphasizes predefined roadway scenarios to categorize risk. Kutela  {\em et al.} \cite{kutela2024influence} leverage overspeed data and roadway features, while Roi Naveiro  {\em et al.} \cite{naveiro2024adversarial} adopt game‐theoretic approaches to lane‐changing. Chen  {\em et al.} \cite{chen2023follownet} focus on data‐driven car‐following behavior, and Cheng  {\em et al.} \cite{cheng2024research} quantify safety risk by analyzing driver cognitive load in high‐risk conditions like distracted driving or lane merging. These efforts underscore the importance of scenario or lane environments but seldom merge multiple core factors—angle, speed, vehicle motion—into a generic, holistic collision‐risk index.
\vspace{-2pt}  

Beyond these complexity or scenario‐based methods, risk‐bounded envelopes have also emerged. Bernhard  {\em et al.} \cite{bernhard2022risk} propose a probabilistic framework for uncertain perception, and the Responsibility-Sensitive Safety (RSS) model \cite{shalev2017formal} applies conservative braking rules. While robust, such envelope‐oriented approaches often overlook subtler environment complexities or the interplay of ego motion and lateral/longitudinal risk. Large‐scale Naturalistic Field Operational Test (N-FOT) studies (e.g., Othman  {\em et al.} \cite{othman2012using}) require substantial data and seldom capture extreme events in detail.
\vspace{-2pt}  

Overall, prior research addresses angles, environment complexity, or safety envelopes separately, indicating a gap for an integrated collision‐risk measure. Inspired by angle‐based collision components \cite{montewka2021collision}, environment complexity \cite{yu2021complexity}, and envelope logic \cite{bernhard2022risk,shalev2017formal}, our CRI seeks to bridge these facets by unifying lateral/longitudinal risk, overspeed, and angle into a lightweight real‐time index.

\section{Methodology}
\label{sec:method}

\begin{figure*}[t]
    \centering
    \includegraphics[width=1\linewidth]{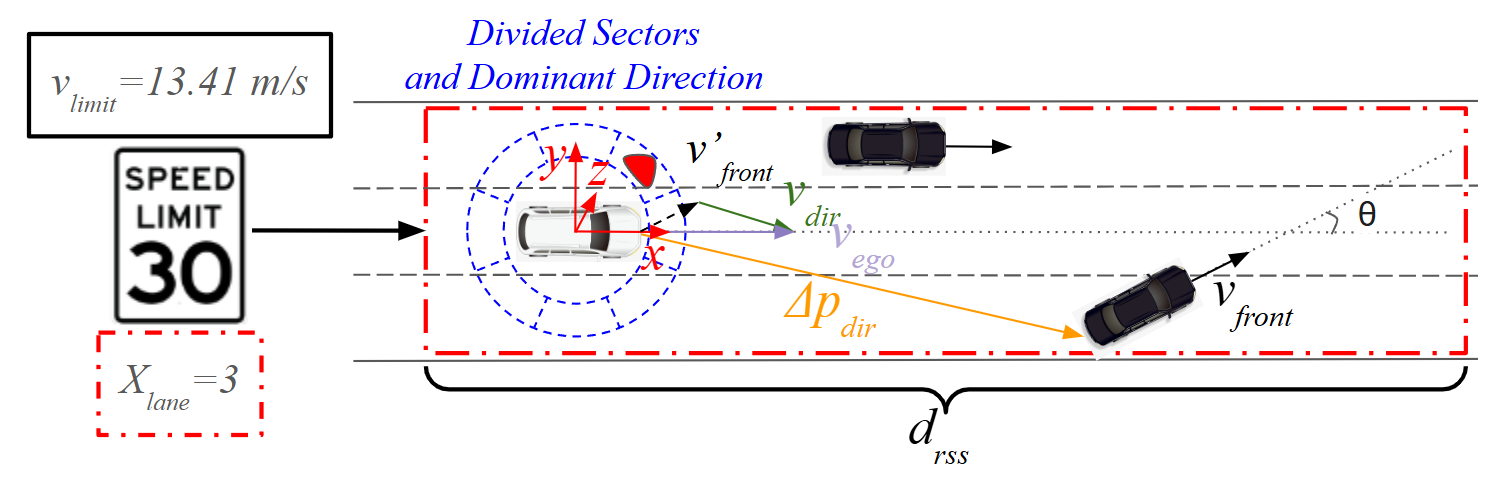}
    \caption{Required inputs for CRI calculation. The framework leverages ego vehicle dynamics (e.g., $v_{ego}$), road parameters (e.g., speed limit $v_{limit}$, number of lanes $X_{lane}$), object states (relative position $\Delta p_{dir}$, relative velocity $v_{dir}$, object velocity $v_{front}$), and directional relationships (e.g., heading angle $\theta$), within an RSS-defined safety envelope $d_{rss}$.}
    \label{fig:needed_info}
\end{figure*}

The Context-aware Risk Index (CRI) enhances autonomous driving safety by integrating vehicle dynamics, probabilistic risk modeling, and safety theory. As shown in Figure~\ref{fig:system_architecture}, the system processes raw sensor input into adaptive control commands. Detected objects are grouped into eight directional zones based on relative position, enabling fine-grained, context-aware risk evaluation and response. The required CRI inputs are summarized in Figure~\ref{fig:needed_info}.

\subsection{CRI Calculation Pipeline}

For each decision cycle, CRI computation operates at two levels. First, a set of risk factors is independently calculated for each surrounding vehicle relative to the ego vehicle, based on their relative kinematic and geometric states. Subsequently, the computed CRI values for all objects are assigned into one of eight spatial sectors based on their relative bearing angles. Within each sector, the maximum per-object CRI is retained, and a hybrid fusion method consolidates the sectoral risks into a final scene-level CRI score and dominant risk direction, which are then used to guide adaptive control decisions.

We initiate the CRI computation pipeline by constructing a dynamic safety envelope around the ego vehicle based on the Responsibility-Sensitive Safety (RSS) model \cite{shalev2017formal}. The longitudinal range of the envelope is determined by the RSS safe distance, while the lateral width is adjusted according to the number of lanes on the current road. The RSS distance is computed as:

\begin{align}
d_{rss} = v_{ego} \cdot t_{reaction} + \frac{1}{2} a_{max} \cdot t_{reaction}^2 \\ 
+ \frac{(v_{ego} + t_{reaction} \cdot a_{max})^2 - v_{front}^2}{2a_{min}},
\end{align}

where \( v_{ego} \) is the ego vehicle's speed, \( v_{front} \) is the leading object's speed, \( t_{reaction} \) is the reaction time, and \( a_{max}, a_{min} \) are the maximum acceleration and minimum braking deceleration.

Next, within the safety envelope, individual risk factors based on kinetic attributes of detected objects are computed for each detected vehicle. Following Montewka  {\em et al.} \cite{montewka2021collision}, we define an orientation risk \( f_{orientation} \) as:

\begin{align}
f_{orientation} = 0.5 \left(1 - \cos\left(\frac{\theta \cdot \pi}{101.25} + \frac{\pi}{10}\right)\right),
\end{align}

where \( \theta \) is the angle between the object's orientation and the ego vehicle’s heading.

We also evaluate risks associated with relative velocities using an \emph{approach check} in the ego-fixed frame. Let \(\Delta p_{dir}\) be the relative positions' projections on longitudinal or latitudinal directions (x or y axis), and \(v_{dir}\) be relative velocities' projections. We define the \emph{approach check}:

\begin{align}
p_{\mathrm{dir}} = \Delta p_{\mathrm{dir}}\;v_{\mathrm{dir}},\quad
\mathrm{dir}\in\{\mathrm{lon},\,\mathrm{lat}\}.
\end{align}

A negative projection (\(p_{\mathrm{dir}}<0\)) means \(\Delta p_{\mathrm{dir}}\) and \(v_{\mathrm{dir}}\) have opposite signs—separation along that axis is decreasing—so the object is approaching; nonnegative (\(p_{\mathrm{dir}}\ge0\)) implies receding and yields zero risk. 

We then define:

\begin{align}
f_{\mathrm{dir}} =
\begin{cases}
0, & p_{\mathrm{dir}} \geq 0,\\
\exp\left( -\mathrm{TTC}_{\mathrm{dir}} \right), & p_{\mathrm{dir}} < 0,
\end{cases} \\
\qquad 
\mathrm{TTC}_{\mathrm{dir}} = \frac{|\Delta p_{\mathrm{dir}}|}{|v_{\mathrm{dir}}| + \epsilon},
\end{align}

where \(\mathrm{TTC}_{\mathrm{dir}})\) (directional Time-to-Collision) represents the time remaining until potential collision on certain directiol, and \(\mathrm{dir}\in\{\mathrm{lon},\mathrm{lat}\}\), \(\epsilon\) is a small constant for stability, and \(\Delta p_{\mathrm{dir}}\) and \(v_{\mathrm{dir}}\) are the corresponding directional projections of relative position and velocity. Objects with nonnegative \(p_{\mathrm{dir}}\) are receding and thus contribute zero to the risk. We adopt exponential decay on TTC to match human risk perception, providing higher sensitivity to imminent collisions.

Inspired by Kutela  {\em et al.} \cite{kutela2024influence}, we also introduce a logistic function integrating the ratio of speed to speed limit and number of lanes, resulting in a more explanatory speed-related risk factor

\begin{align}
f_{speed} = \frac{1}{1 + e^{-(5\frac{v_{ego}-v_{limit}}{v_{limit}} + 1.5X_{lane} - 2)}},
\end{align}

where \( v_{limit} \) is the speed limit and \( X_{lane} \) adjusts for the number of lanes. The constants are empirically determined based on \cite{kutela2024influence} to balance speed penalty and lane safety considerations. And the logistic function's non-linearity accurately models the rapid increase in collision severity with velocity. This effectively penalizes higher speeds, encouraging conservative driving. 

These computed risk factors are fused into a unified CRI using a probabilistic-max hybrid strategy. We adopt this probabilistic-max hybrid strategy to balance cumulative risk aggregation with peak threat prioritization, ensuring both comprehensive risk coverage and sensitivity to extreme scenarios. Each risk component \( f_i \in \{f_{\text{orientation}}, f_{\text{lon}}, f_{\text{lat}} \} \) naturally falls within the \([0,1]\) range due to their mathematical formulation. Specifically, \( f_{\text{orientation}} \) is derived from a bounded trigonometric function, while \( f_{\text{lon}} \) and \( f_{\text{lat}} \) are computed using exponential decay based on time-to-collision, ensuring all components remain comparable for reliable fusion.  Spatial risk \( f_{\text{spatial}} \) is defined as:

\begin{align}
f_{\text{spatial}} = 1 - \prod_{i}(1 - f_i), \quad f_i \in \{f_{\text{orientation}}, f_{\text{lon}}, f_{\text{lat}}\}.
\end{align}

To ensure that extreme risk scenarios are properly prioritized, we compute the maximum individual risk factor across all evaluated dimensions, denoted as \(\max_i f_i\). Following prospect theory \cite{Kahneman_Tversky_1979}, which emphasizes the human tendency to overweight extreme outcomes, we blend the cumulative spatial risk \( f_{\text{spatial}} \) and the maximum single risk \(\max_i f_i\) using a weighting coefficient \( \alpha = 0.7 \):

\begin{align}
\text{CRI} = \left[ \alpha \cdot f_{\text{spatial}} + (1-\alpha) \cdot \max_i f_i \right] \cdot \frac{e^{f_{speed} - \text{speed\_ref}}}{e^{\text{speed\_ref}}},
\end{align}

where \( f_{\text{speed}} \in [0,1] \) reflects the ego vehicle's speed-related risk, and \( \text{speed\_ref} \) is a calibrated neutrality point. The exponential modulation ensures that CRI values are further amplified at higher ego speeds, reflecting the greater severity of potential collisions under elevated kinetic energy.

At this stage, each surrounding vehicle has an associated CRI value, reflecting its directional threat, proximity risk, and dynamic severity relative to the ego vehicle. To construct a global view of the surrounding environment, these individual CRI scores are then spatially organized based on their relative bearing angles.

\subsection{Direction-aware CRI Aggregation}

To enable directional risk awareness, we divide the surrounding space into eight equally spaced sectors around the ego vehicle, aligned with its body-fixed frame. This sectorization ensures that threats from different directions are separately preserved, supporting precise and context-sensitive control decisions.

For each vehicle within the RSS-defined safety envelope, its computed CRI is assigned to the corresponding sector based on relative bearing, and within each sector \( d \), we retain the maximum CRI value \( R_d \) across all contributing objects.

To consolidate the directional risk information, we perform a hybrid vector-max fusion. Specifically, we compute an aggregated risk vector magnitude \( R_{\text{vector}} \) as:
\begin{align}
R_{\text{vector}} = \sqrt{\left(\sum_{d} R_d \cos \theta_d\right)^2 + \left(\sum_{d} R_d \sin \theta_d\right)^2},
\end{align}
where \( \theta_d \) is the central angle of sector \( d \). In parallel, the maximum directional risk is identified as \( R_{\text{max}} = \max_d R_d \).

The final aggregated CRI score is obtained through a weighted sum:
\begin{align}
\text{CRI}_{\text{final}} = \beta R_{\text{vector}} + (1-\beta) R_{\text{max}},   
\end{align}
where \( \beta = 0.7 \) balances between smooth spatial integration and peak risk sensitivity.

Finally, to guide control adaptation, we identify the \emph{dominant risk direction} as the sector \( d^* \) with the highest \( R_d \), with corresponding angle \(\theta^* = \theta_{d^*}\). This real-time directional index informs throttle, brake, and steering adjustments based on the location of the most critical threat.

\begin{figure}[t]
\centering
\includegraphics[width=0.5\textwidth]{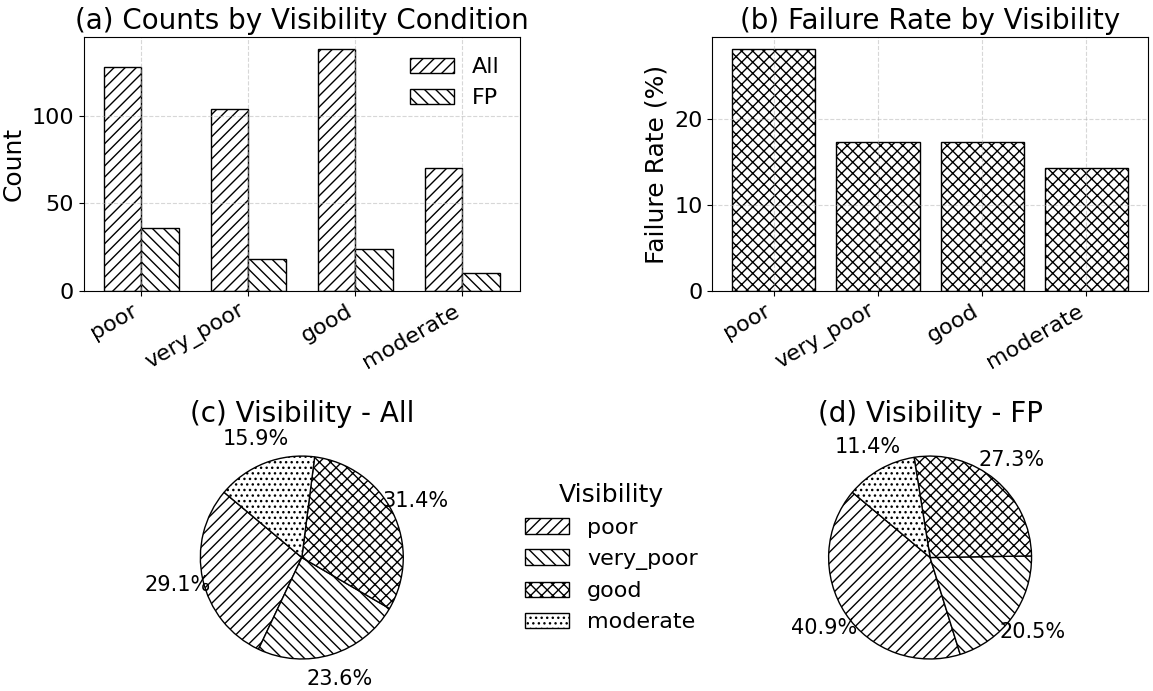}
\caption{Visibility Condition Analysis across the Bench2Drive dataset. (a) Counts of all routes and failure routes by visibility; (b) Failure rates under different visibility conditions; (c) Visibility distribution in all routes; (d) Visibility distribution in failure-prone routes.}
\label{fig:dataset_distribution}
\end{figure}

\section{Experiments}

We evaluate the integration of the Context-aware Risk Index (CRI) with a state-of-the-art end-to-end autonomous driving model Transfuser++ by Jaeger  {\em et al.} \cite{Jaeger_Chitta_Geiger_2023} within the CARLA simulator \cite{Dosovitskiy17}. Testing is performed on the Bench2Drive dataset \cite{Jia_Yang_Li_Zhang_Yan_2024}, which offers challenging, diverse scenarios specifically crafted to expose system weaknesses under safety-critical conditions.

\subsection{Experimental Setup}

We evaluate the Context-aware Risk Index (CRI) framework by integrating it into Transfuser++, currently the strongest open-source end-to-end autonomous driving model on CARLA Leaderboard. Experiments are conducted on the Bench2Drive dataset, comprising 220 safety-critical scenarios generated in the CARLA simulator. Bench2Drive systematically covers extreme traffic behaviors, such as dynamic obstacles and intersection violations under varied environmental conditions. It focuses on exposing decision-making vulnerabilities rather than sensor-level failures, providing a comprehensive benchmark for safety-critical evaluation. Visibility conditions are distributed across poor (29.1\%), very poor (23.6\%), good (31.4\%), and moderate (15.9\%) levels (Figure~\ref{fig:dataset_distribution}(c)). 

To further stress-test CRI, we isolate 44 failure-prone scenarios (FP) where the baseline agent experienced collisions. Poor and very poor visibility cases are overrepresented (40.9\% and 20.5\%, (Figure~\ref{fig:dataset_distribution}(c))), and failures occur even under good visibility (Figure~\ref{fig:dataset_distribution}(a)(b)), highlighting the depth of decision-making challenges captured in the FP subset.

The CRI module is integrated into the CARLA Leaderboard evaluation framework with minimal modifications, focusing exclusively on control-level adaptation. All tests are performed in synchronous mode under the \texttt{SENSORS} benchmark track, with results logged per route.

\subsection{Adaptive Control Policy}

The CRI module enhances safety by dynamically adapting the vehicle's control behavior in response to real-time directional risk assessments. We predefine three distinct autonomous driving styles: \textit{aggressive}, \textit{neutral}, and \textit{conservative}, each corresponding to different control sensitivity levels. Based on the aggregated and directional CRI values, the system selects the appropriate driving mode at each timestep.
\begin{figure*}[t]
    \centering
    \includegraphics[width=1\textwidth]{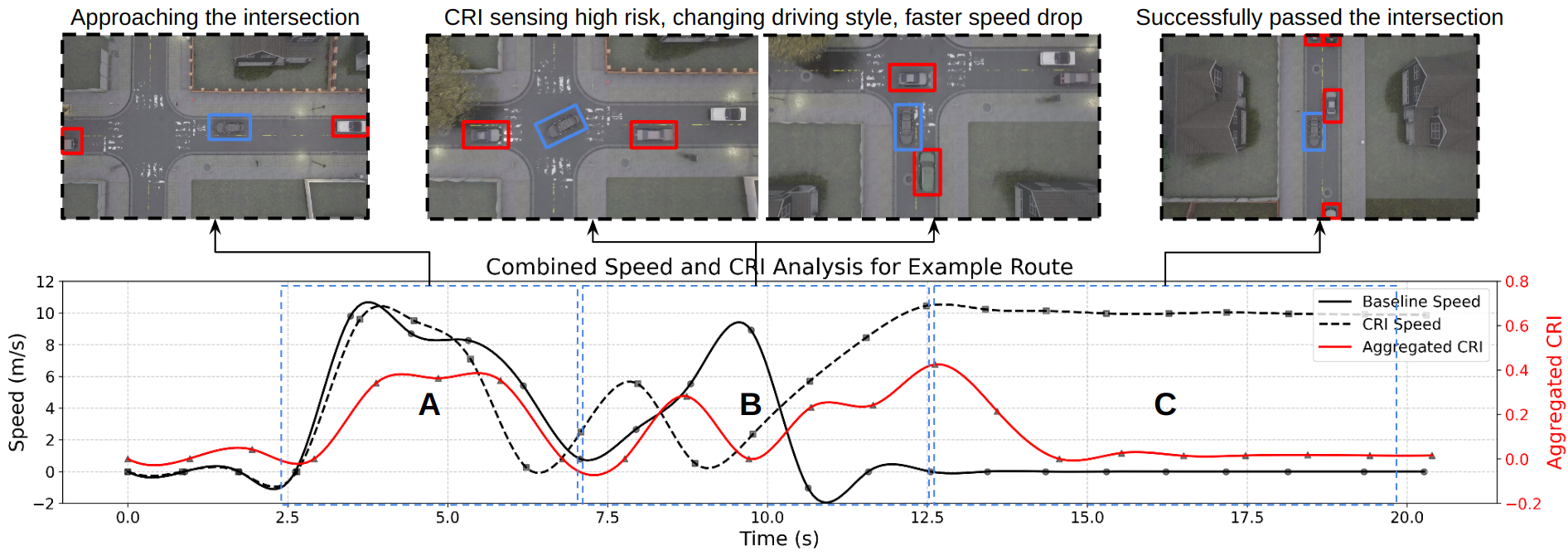}
    \caption{Demonstration of CRI-enabled hazard mitigation during an extreme four-way intersection encounter. The ego vehicle encounters an opposing vehicle violating a stop sign. CRI enables earlier risk detection, stronger braking responses, and conservative post-hazard behavior, successfully avoiding collision compared to baseline control.}
    \label{fig:example_route}
\end{figure*}

The adaptive control logic proceeds as follows:

\begin{algorithm}[H]
\caption{CRI-Based Adaptive Control Policy}
\begin{algorithmic}[1]
\State \textbf{Input:} Ego state $s_{ego}$, detected objects $\mathcal{O} = \{o_i\}$
\State \textbf{Initialize:} Risk calculator $\mathcal{C}$, adaptive controller $\mathcal{A}$
\While{navigation is ongoing}
    \State Update object list $\mathcal{O}$
    \For{each $o_i \in \mathcal{O}$}
        \State Compute directional risk $r_i = \mathcal{C}(o_i)$
    \EndFor
    \State Aggregate risks into distribution vector $\mathbf{r}$
    \State Identify dominant risk direction $\theta^*$ and risk magnitude $r^*$
    \State Select driving mode $m \leftarrow \mathcal{M}(r^*, \theta^*)$
    \State Compute adapted control $u_{control} = \mathcal{A}(s_{ego}, m)$
    \State \textbf{Return} $u_{control}$
\EndWhile
\end{algorithmic}
\end{algorithm}

This design highlights the modularity and real-time adaptability of CRI: it introduces no additional retraining overhead and provides a flexible, principled mechanism for safety-enhancing adjustments in dynamic driving environments.

\section{Evaluation}

We evaluate CRI’s effectiveness in improving safety, comfort, and runtime efficiency across diverse high-risk scenarios. Experiments cover both failure-prone and general cases, and include analysis of runtime overhead and system limitations.

\subsection{Illustrative Case Study: Emergency Handling at Intersection}
We first illustrate the comprehensive working mechanism of CRI through a carefully selected representative extreme scenario from the Bench2Drive dataset that exemplifies the type of safety-critical situations where traditional autonomous driving systems often struggle. In this particularly challenging test case, the ego vehicle must successfully navigate through a complex four-way intersection while simultaneously dealing with an opposing vehicle that illegally runs a stop sign, thereby creating a highly hazardous and unpredictable situation that closely mirrors the type of dangerous traffic violations commonly encountered in real-world driving scenarios and frequently cited in traffic accident reports. Figure~\ref{fig:example_route} provides a comprehensive visualization showing the complete sequence of events alongside detailed comparisons of the ego vehicle's speed profiles under both Baseline and CRI-enabled control methods, as well as the corresponding aggregated CRI signal evolution over the entire duration of the encounter.

\begin{table*}[h]
\captionsetup{belowskip=0pt}
\centering
\setlength{\tabcolsep}{16pt}
\small
\caption{Comparison of Baseline and CRI Performance on Failure-Prone and All Bench2Drive Scenarios.}
\label{tab:merged_comparison}
\begin{tabular}{lcccccccc}
\toprule
 & \textbf{CpR} & \textbf{CpK} & \textbf{CS} & \textbf{SP} & \textbf{RT} & \textbf{MAJ} & \textbf{SAJ} & \textbf{MJ} \\
\midrule
Baseline\_FP & 1.23 & 12.37 & 44.96 & 0.46 & 0.05 & 33.34 & 95.55 & 673.08 \\
CRI\_FP & 1.00 & 9.92 & 52.62 & 0.53 & 0.02 & 27.09 & 68.16 & 562.82 \\
Baseline\_ALL & 0.38 & 2.99 & 80.06 & 0.20 & 0.04 & 28.46 & 55.98 & 327.69 \\
CRI\_ALL & 0.32 & 2.42 & 83.19 & 0.19 & 0.03 & 22.73 & 44.36 & 300.03 \\
\bottomrule
\end{tabular}
\captionsetup{justification=centering}
\caption*{\small \textbf{Abbreviations:} CpR = Collisions per Route; CpK = Collisions per Kilometer; CS = Composed Score; SP = Score Penalty; RT = Route Timeout Rate; MAJ = Mean Absolute Jerk; SAJ = Std Absolute Jerk; MJ = Max Jerk. FP = Failure-Prone Scenarios; ALL = All Bench2Drive Scenarios.}
\end{table*}

In stage \textbf{A} (approaching the intersection), the ego vehicle begins its deceleration process as it approaches the stop sign in preparation for the intersection crossing. During this critical approach phase, the presence of a following vehicle that is closing in on the ego vehicle's position triggers an increase in the CRI value, which effectively captures the developing risk situation through its directional risk assessment capabilities. This rising CRI signal prompts the adaptive controller to implement more aggressive braking input compared to the baseline approach, demonstrating the system's proactive risk mitigation strategy. As a direct result of this enhanced risk-aware control intervention, the ego vehicle achieves a more stable and controlled deceleration profile, ultimately coming to a complete stop both earlier in time and with greater security margins than the baseline method, thereby establishing a safer starting position for the subsequent intersection crossing maneuver.

In stage \textbf{B} (crossing triggered and hazard emergence), as the ego vehicle begins to proceed forward into the intersection area, the opposing vehicle suddenly and unexpectedly violates the stop sign traffic regulation, creating an immediate and severe collision threat. The CRI system rapidly detects this high-risk situation through its real-time directional risk monitoring capabilities, resulting in a sharp local peak in the CRI signal that accurately reflects the sudden escalation in danger level. In direct response to this elevated risk assessment, the adaptive controller immediately commands a stronger and more decisive braking action, which leads to a rapid and controlled drop in the ego vehicle's speed to avoid the imminent collision. The baseline agent, operating without the benefit of CRI's enhanced risk awareness capabilities, demonstrates significantly slower reaction times and inadequate response measures, ultimately resulting in a collision with the opposing vehicle due to insufficient braking force and delayed threat recognition. In stark contrast, the CRI-enabled vehicle successfully waits for the opposing vehicle to complete its illegal maneuver and pass through the intersection before cautiously resuming its own forward motion. The presence of lingering elevated CRI values during this recovery phase ensures that the ego vehicle accelerates in a conservative manner, effectively reflecting the system's heightened vigilance and continued awareness of the residual threats posed by nearby traffic participants who may exhibit unpredictable behavior patterns.

Finally, in stage \textbf{C} (exiting the intersection), after the ego vehicle has successfully cleared the hazardous intersection zone and moved beyond the immediate threat area, the CRI signal gradually returns to near-zero levels, indicating that the system has determined the risk environment has returned to normal baseline conditions, and consequently the vehicle resumes its normal driving behavior patterns and speed profiles.

This case study demonstrates that CRI not only senses dynamic risk conditions earlier but also translates risk perception into timely, safety-aware control adaptations, allowing the ego vehicle to avoid collisions even under sudden extreme hazards.

\subsection{Performance on Failure-Prone Scenarios}

We first evaluate CRI's effectiveness on the carefully selected subset of 44 failure-prone (FP) scenarios where baseline agents consistently exhibited collisions, representing the most challenging and safety-critical situations within the Bench2Drive benchmark that expose fundamental limitations in conventional autonomous driving approaches. CRI demonstrates consistent and statistically significant improvements across all key safety and comfort metrics, indicating robust performance enhancement specifically in the scenarios where it is most critically needed. 

The average number of vehicle collisions per route exhibits a substantial decrease from 1.23 to 1.00, representing an 18.7\% reduction with strong statistical significance (p = 0.003), while the collision rate per kilometer shows an even more pronounced improvement, dropping from 12.37 to 9.92 collisions per kilometer, which corresponds to a 19.8\% reduction that is highly statistically significant (p = 0.004). These collision reduction results provide compelling evidence of CRI's fundamental capability to recognize imminent threats through its directional risk assessment mechanisms and dynamically adjust control strategies in real-time to prevent accidents that would otherwise be unavoidable using conventional approaches. Beyond pure safety metrics, the composed driving score demonstrates substantial improvement from 44.96 to 52.62, representing a significant 17.0\% increase (p = 0.016) that reflects not only collision avoidance but also enhanced overall task completion and adherence to traffic rules, while the score penalty metric shows favorable enhancement from 0.46 to 0.53, indicating a 15.2\% increase (p = 0.013) that directly reflects fewer rule violations and better task completion outcomes throughout the challenging scenario sequences. In terms of driving comfort and passenger experience, CRI delivers particularly impressive improvements in vehicle dynamics smoothness, with the standard deviation of jerk decreasing dramatically from 95.55 to 68.16 m/s³, representing a substantial 28.7\% reduction that achieves statistical significance (p = 0.021) and directly indicates much smoother vehicle dynamics and more pleasant riding experiences. This significant comfort improvement demonstrates that CRI effectively reduces unnecessary accelerations and braking fluctuations, thereby leading to enhanced ride quality that benefits both safety and passenger comfort simultaneously. These comprehensive results validate CRI's fundamental strength in providing effective real-time risk mitigation capabilities specifically under highly adversarial conditions where conventional autonomous driving systems struggle most significantly.

\subsection{Performance across All Bench2Drive Scenarios}

We further assess CRI across all 220 Bench2Drive scenarios, encompassing diverse high-risk situations. As illustrated in Figure~\ref{fig:dataset_distribution}, while all scenarios are extreme by design, the failure-prone (FP) subset features a higher concentration of poor visibility conditions, making it inherently more challenging.

Across all scenarios, CRI reduces the average vehicle collisions per route from 0.38 to 0.32 (15.8\% reduction, p = 0.007) and the collision rate per kilometer from 2.99 to 2.42 (19.1\% reduction, p = 0.008), confirming robust performance even beyond the highest-risk cases. The composed driving score increases from 80.06 to 83.19 (3.9\% gain, p = 0.038), slightly smaller than in the FP subset, reflecting the broader diversity of difficulty levels. Comfort metrics also improve: the mean absolute jerk decreases from 28.46 to 22.73 m/s³ (20.2\% reduction, p = 0.042), and the jerk standard deviation drops from 55.98 to 44.36 m/s³ (20.8\% reduction, p = 0.045), confirming that CRI enhances control stability consistently across diverse high-risk scenarios.

Meanwhile, the score penalty slightly decreases from 0.20 to 0.19 (5.0\% reduction, p = 0.312), though the change is not statistically significant. This mild decrease likely reflects that, in the full dataset, many routes are completed with relatively few infractions even under baseline control, limiting the room for further improvement by CRI.

Overall, these results confirm that CRI not only improves safety and comfort under the most failure-prone conditions but also provides systematic gains across a comprehensive range of extreme driving tasks, validating its generalizability as a lightweight risk adaptation layer.

\subsection{System Modularity and Runtime Efficiency}

CRI is designed with modularity and efficiency in mind. Risk estimation and control adaptation are separated into independent modules with simple interfaces, allowing seamless integration into existing driving agents with minimal code changes.

Runtime profiling shows that CRI introduces negligible overhead. As summarized in Table~\ref{tab:cri_runtime}, CRI adds only 3.60 ms per decision cycle on average, corresponding to a 5.10\% increase over the baseline runtime (74.13 ms vs. 70.53 ms). The risk calculation and adaptation steps each require less than 0.03 ms per cycle.

\begin{table}[h]
\centering
\small
\caption{CRI Runtime Overhead Analysis}
\label{tab:cri_runtime}
\begin{tabular}{lc}
\toprule
\textbf{Component} & \textbf{Mean Runtime (ms)} \\
\midrule
CRI Reception & 0.020 \\
Control Adaptation & 0.006 \\
Controller Initialization & 0.005 \\
\midrule
RunStep (Baseline) & 70.53 \\
RunStep (with CRI) & 74.13 \\
Overhead & 3.60 \\
Overhead (\%) & 5.10\% \\
\bottomrule
\end{tabular}
\end{table}

\section{Limitations, Discussion, and Future Work}

While CRI brings substantial improvements in safety and comfort, some limitations persist. The composed driving score increases modestly by 3.9\%, and the score penalty shows only minor improvement, reflecting that many scenarios are already handled well by the baseline. In addition, current evaluations are limited to a single end-to-end backbone in simulation; broader testing across architectures, real-world datasets, and physical vehicles is needed to fully confirm generalizability. Future work will address these challenges and incorporate machine learning techniques to optimize CRI parameters dynamically. We also aim to explore integration of CRI signals into perception and planning modules, balance safety, comfort, and efficiency across diverse operational contexts. 

\section*{Acknowledgement}
This work is in part supported by the US National Science Foundation under Award No. NSF~\#2416937.


\bibliographystyle{IEEEtran}

\begin{thebibliography}{10}
\providecommand{\url}[1]{#1}
\csname url@samestyle\endcsname
\providecommand{\newblock}{\relax}
\providecommand{\bibinfo}[2]{#2}
\providecommand{\BIBentrySTDinterwordspacing}{\spaceskip=0pt\relax}
\providecommand{\BIBentryALTinterwordstretchfactor}{4}
\providecommand{\BIBentryALTinterwordspacing}{\spaceskip=\fontdimen2\font plus
\BIBentryALTinterwordstretchfactor\fontdimen3\font minus \fontdimen4\font\relax}
\providecommand{\BIBforeignlanguage}[2]{{%
\expandafter\ifx\csname l@#1\endcsname\relax
\typeout{** WARNING: IEEEtran.bst: No hyphenation pattern has been}%
\typeout{** loaded for the language `#1'. Using the pattern for}%
\typeout{** the default language instead.}%
\else
\language=\csname l@#1\endcsname
\fi
#2}}
\providecommand{\BIBdecl}{\relax}
\BIBdecl

\bibitem{yu2021complexity}
R.~Yu, Y.~Zheng, and X.~Qu, ``Dynamic driving environment complexity quantification method and its verification,'' \emph{Transportation Research Part C}, vol. 127, p. 103051, 2021.

\bibitem{liu2019towards}
Y.~Liu and J.~H. Hansen, ``Towards complexity level classification of driving scenarios using environmental information,'' in \emph{2019 IEEE Intelligent Transportation Systems Conference (ITSC)}.\hskip 1em plus 0.5em minus 0.4em\relax IEEE, 2019, pp. 810--815.

\bibitem{kutela2024influence}
B.~Kutela, F.~Ngeni, C.~Ruseruka, T.~J. Chengula, N.~Novat, H.~Shita, and A.~Kinero, ``The influence of roadway characteristics and built environment on the extent of over-speeding: An exploration using mobile automated traffic camera data,'' \emph{International Journal of Transportation Science and Technology}, 2024.

\bibitem{naveiro2024adversarial}
R.~Naveiro, D.~R{\'\i}os~Insua, and W.~N. Caballero, ``Adversarial risk analysis for automated lane-changing in heterogeneous traffic,'' in \emph{International Conference on Algorithmic Decision Theory}.\hskip 1em plus 0.5em minus 0.4em\relax Springer, 2024, pp. 128--143.

\bibitem{Jaeger_Chitta_Geiger_2023}
\BIBentryALTinterwordspacing
B.~Jaeger, K.~Chitta, and A.~Geiger, ``Hidden biases of end-to-end driving models,'' no. arXiv:2306.07957, Aug. 2023, arXiv:2306.07957 [cs]. [Online]. Available: \url{http://arxiv.org/abs/2306.07957}
\BIBentrySTDinterwordspacing

\bibitem{Jia_Yang_Li_Zhang_Yan_2024}
\BIBentryALTinterwordspacing
X.~Jia, Z.~Yang, Q.~Li, Z.~Zhang, and J.~Yan, ``Bench2drive: Towards multi-ability benchmarking of closed-loop end-to-end autonomous driving,'' no. arXiv:2406.03877, Nov. 2024, arXiv:2406.03877 [cs]. [Online]. Available: \url{http://arxiv.org/abs/2406.03877}
\BIBentrySTDinterwordspacing

\bibitem{Dosovitskiy17}
A.~Dosovitskiy, G.~Ros, F.~Codevilla, A.~Lopez, and V.~Koltun, ``{CARLA}: {An} open urban driving simulator,'' in \emph{Proceedings of the 1st Annual Conference on Robot Learning}, 2017, pp. 1--16.

\bibitem{montewka2021collision}
J.~Montewka, F.~Goerlandt, and P.~Kujala, ``Determination of collision criteria and causation factors appropriate to a model for estimating the probability of maritime accidents,'' \emph{Ocean Engineering}, vol.~40, pp. 50--61, 2021.

\bibitem{gold2016taking}
C.~Gold, M.~K{\"o}rber, D.~Lechner, and K.~Bengler, ``Taking over control from highly automated vehicles in complex traffic situations: The role of traffic density,'' \emph{Human factors}, vol.~58, no.~4, pp. 642--652, 2016.

\bibitem{faure2016effects}
V.~Faure, R.~Lobjois, and N.~Benguigui, ``The effects of driving environment complexity and dual tasking on drivers’ mental workload and eye blink behavior,'' \emph{Transportation research part F: traffic psychology and behaviour}, vol.~40, pp. 78--90, 2016.

\bibitem{guo2020safedata}
J.~Guo, U.~Kurup, and M.~Shah, ``Is it safe to drive? an overview of factors, metrics, and datasets for driveability assessment in autonomous driving,'' \emph{IEEE Transactions on Intelligent Transportation Systems}, vol.~21, no.~8, pp. 3135--3151, 2020.

\bibitem{wang2018traffic}
J.~Wang, C.~Zhang, Y.~Liu, and Q.~Zhang, ``Traffic sensory data classification by quantifying scenario complexity,'' in \emph{2018 IEEE Intelligent Vehicles Symposium (IV)}.\hskip 1em plus 0.5em minus 0.4em\relax IEEE, 2018, pp. 1543--1548.

\bibitem{manawadu2018estimating}
U.~E. Manawadu, T.~Kawano, S.~Murata, M.~Kamezaki, and S.~Sugano, ``Estimating driver workload with systematically varying traffic complexity using machine learning: Experimental design,'' in \emph{IHSI 2018: Integrating People and Intelligent Systems, January 7-9, 2018, Dubai, United Arab Emirates}.\hskip 1em plus 0.5em minus 0.4em\relax Springer, 2018, pp. 106--111.

\bibitem{feng2023dense}
S.~Feng, H.~Sun, X.~Yan, H.~Zhu, Z.~Zou, S.~Shen, and H.~X. Liu, ``Dense reinforcement learning for safety validation of autonomous vehicles,'' \emph{Nature}, vol. 615, no. 7953, pp. 620--627, 2023.

\bibitem{garefalakis2022data}
T.~Garefalakis, C.~Katrakazas, and G.~Yannis, ``Data-driven estimation of a driving safety tolerance zone using imbalanced machine learning,'' \emph{Sensors}, vol.~22, no.~14, p. 5309, 2022.

\bibitem{chen2023follownet}
X.~Chen, M.~Zhu, K.~Chen, P.~Wang, H.~Lu, H.~Zhong, X.~Han, X.~Wang, and Y.~Wang, ``Follownet: a comprehensive benchmark for car-following behavior modeling,'' \emph{Scientific data}, vol.~10, no.~1, p. 828, 2023.

\bibitem{cheng2024research}
L.~Cheng, H.~Ren, H.~Guo, and D.~Cao, ``Research on the evaluation method for safety cognitive ability of workers in high-risk construction positions,'' \emph{Engineering, Construction and Architectural Management}, 2024.

\bibitem{bernhard2022risk}
J.~Bernhard, P.~Hart, A.~Sahu, C.~Sch{\"o}ller, and M.~G. Cancimance, ``Risk-based safety envelopes for autonomous vehicles under perception uncertainty,'' in \emph{2022 IEEE Intelligent Vehicles Symposium (IV)}.\hskip 1em plus 0.5em minus 0.4em\relax IEEE, 2022, pp. 104--111.

\bibitem{shalev2017formal}
S.~Shalev-Shwartz, S.~Shammah, and A.~Shashua, ``On a formal model of safe and scalable self-driving cars,'' \emph{arXiv preprint arXiv:1708.06374}, 2017.

\bibitem{othman2012using}
S.~Othman, R.~Thomson, and G.~Lann{\'e}r, ``Using naturalistic field operational test data to identify horizontal curves,'' \emph{Journal of transportation engineering}, vol. 138, no.~9, pp. 1151--1160, 2012.

\bibitem{Kahneman_Tversky_1979}
D.~Kahneman and A.~Tversky, ``Prospect theory: An analysis of decision under risk,'' \emph{Econometrica}, vol.~47, no.~2, p. 263–291, 1979.

\end{thebibliography}

\end{document}